\documentclass[10pt,twocolumn,letterpaper]{article}

\usepackage[pagenumbers]{cvpr} % To force page numbers, e.g. for an arXiv version
\usepackage[accsupp]{axessibility}  % Improves PDF readability for those with disabilities.

\usepackage{lipsum}
\usepackage{array}
\usepackage{times}
\usepackage{epsfig}
\usepackage{graphicx}
\usepackage{float}
\usepackage{wrapfig}
\usepackage{amsmath,amssymb,amsthm}
\usepackage{algorithm,algorithmicx,algpseudocode}
\usepackage{bm,xspace}
\usepackage{comment}
\usepackage{multirow}
\usepackage{balance}
\usepackage{url}
\usepackage{booktabs}
\usepackage{etoolbox,siunitx}
\usepackage{calc}
\usepackage{pifont,hologo}
\usepackage{color}
\usepackage{adjustbox}
\usepackage{amsmath}
\usepackage{enumitem}
\usepackage{bbding}  %
\usepackage[normalem]{ulem}  %
\usepackage{contour}
\usepackage{soul}%
\usepackage{tocloft} %
\usepackage{etoc} %
\definecolor{cvprblue}{rgb}{0.21,0.49,0.74}
\usepackage[pagebackref,breaklinks,colorlinks,allcolors=cvprblue]{hyperref}
\usepackage{natbib}
\PassOptionsToPackage{square,numbers,comma,sort}{natbib}

\definecolor{blue}{HTML}{004bb3}
\definecolor{red}{HTML}{cc1100}
\definecolor{orange}{HTML}{cc7700}
\definecolor{gray}{HTML}{efefef}
\definecolor{darkgreen}{HTML}{228B22}
\definecolor{darkgray}{HTML}{757575}

\definecolor{cite}{HTML}{3270b5}
\definecolor{link}{HTML}{b53532}
\definecolor{link}{HTML}{cc1100}
\definecolor{normal}{HTML}{001219}
\definecolor{tta}{HTML}{0077BE}
\definecolor{peft}{HTML}{FFA07A}
\definecolor{seablue}{RGB}{70, 130, 180}
\definecolor{lightorange}{RGB}{255, 180, 140}

\contourlength{0.8pt}

\renewcommand{\eqref}[1]{Eq.~\ref{#1}}

\newcolumntype{x}[1]{>{\centering\arraybackslash}p{#1}}
\newcolumntype{y}[1]{>{\raggedright\arraybackslash}p{#1}}
\newcolumntype{z}[1]{>{\raggedleft\arraybackslash}p{#1}}

\DeclareMathSymbol{@}{\mathord}{letters}{"3B}

\makeatletter
\DeclareRobustCommand\onedot{\futurelet\@let@token\@onedot}
\def\@onedot{\ifx\@let@token.\else.\null\fi\xspace}

\newcommand*{\Rom}[1]{\expandafter\@slowromancap\romannumeral #1@}
\newcommand*{\rom}[1]{\expandafter\romannumeral #1}

\def\1{\bm{1}}

\DeclareMathAlphabet{\mathsfit}{\encodingdefault}{\sfdefault}{m}{sl}
\SetMathAlphabet{\mathsfit}{bold}{\encodingdefault}{\sfdefault}{bx}{n}

\let\originalleft\left
\let\originalright\right
\renewcommand{\left}{\mathopen{}\mathclose\bgroup\originalleft}
\renewcommand{\right}{\aftergroup\egroup\originalright}

\def\paperID{} % *** Enter the Paper ID here
\def\confName{CVPR}
\def\confYear{2026}

\begin{document}

\title{RecycleLoRA: Rank-Revealing QR-Based Dual-LoRA Subspace Adaptation \\ for Domain Generalized Semantic Segmentation}

\author{Chanseul Cho\qquad Seokju Yun\qquad Jeaseong Jeon\qquad Seungjae Moon\qquad  Youngmin Ro\textsuperscript{*} \\Machine Intelligence Laboratory, University of Seoul, Korea \\ \tt\small \{chanseul2001, wsz871, jasonjun1121, msj0243, youngmin.ro\}@uos.ac.kr \\ {\tt\small \url{https://github.com/chanseul01/RecycleLoRA.git}}}

\twocolumn[{%
\renewcommand\twocolumn[1][]{#1}%
\maketitle

}]
\begin{abstract}
Domain Generalized Semantic Segmentation (DGSS) aims to maintain robust performance across unseen target domains. Vision Foundation Models (VFMs) offer rich multi-domain knowledge that can enhance generalization. However, strategies for actively exploiting the rich subspace structures within VFMs remain under-explored, with many existing methods focusing primarily on preserving pre-trained knowledge. Furthermore, their LoRA components often suffer from limited representational diversity and inefficient parameter utilization. We propose RecycleLoRA, which addresses both challenges by employing Rank-Revealing QR Decomposition (RRQR) to systematically exploit VFM’s subspace structures and enhance LoRA’s representational richness. Our main adapter leverages minor subspace directions identified by RRQR to learn diverse and independent features, achieving competitive performance even when used alone. We further introduce a sub adapter that carefully refines major directions with minimal adjustments, providing complementary improvements to the main adapter's strong baseline performance. This design enables the dual adapters to learn distinct representations without requiring additional regularization losses. Our systematic exploitation of pre-trained subspace structures through RRQR-based initialization leads to superior domain generalization performance. RecycleLoRA achieves state-of-the-art performance on both synthetic-to-real generalization and real-to-real generalization tasks without complex architectures or additional inference latency.
\end{abstract}

\vspace{-0.6cm}
\section{Introduction}

Semantic segmentation assigns semantic labels to every pixel in an image and plays a crucial role in autonomous driving, medical imaging, and robotics. However, domain shift, the phenomenon where models trained on one domain experience performance degradation when applied to another, limits real-world deployment. To address this challenge, Domain Generalized Semantic Segmentation (DGSS) has been introduced, aiming to develop models that maintain robust performance across diverse domains without target domain data. This capability is particularly essential in safety-critical applications where models must reliably handle varying conditions. 

\begin{figure}[t]
  \centering
  \includegraphics[width=\columnwidth]{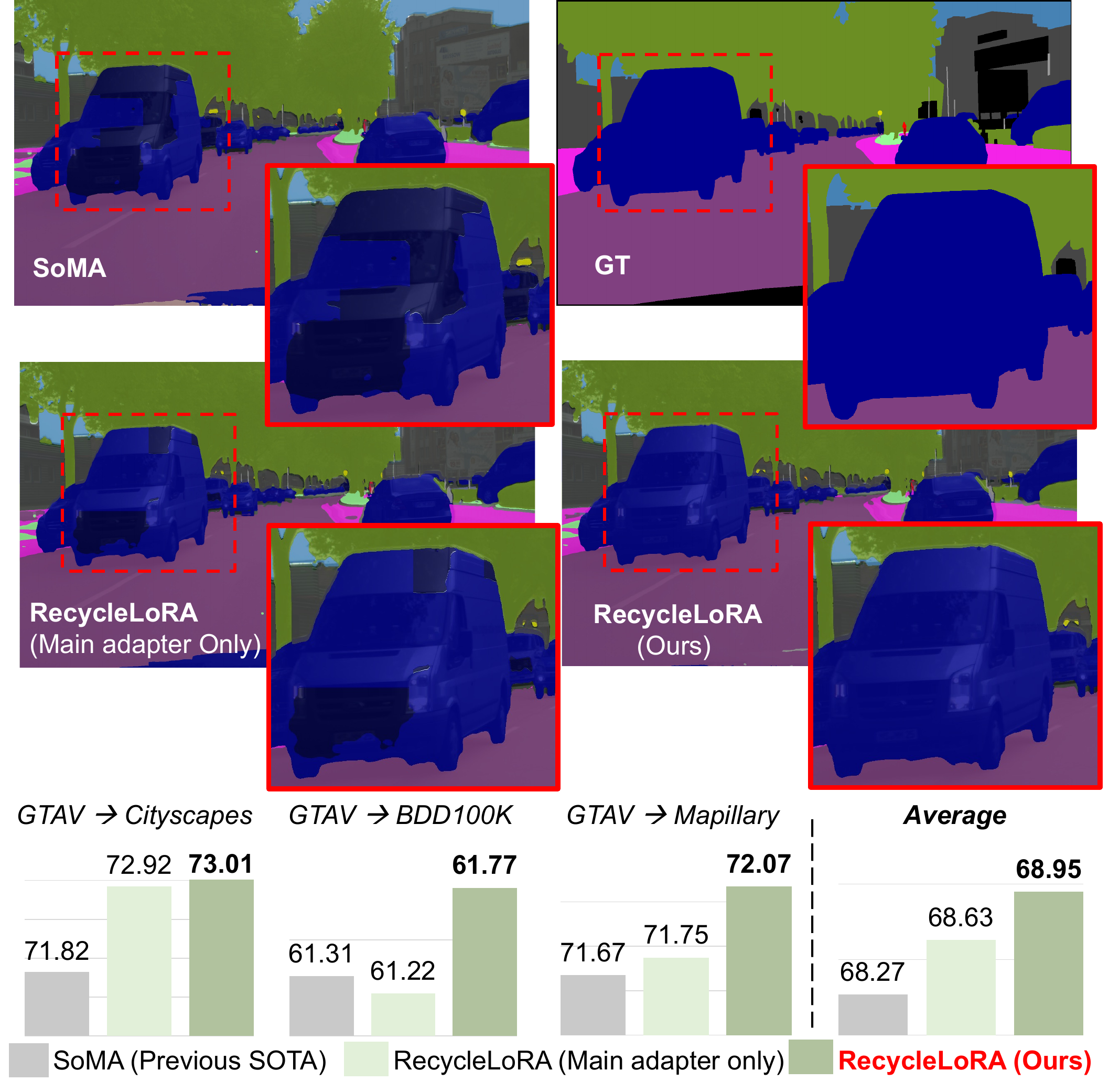}
  \caption{
     Comparison of synthetic-to-real generalization performance (mIoU,\%) between our proposed RecycleLoRA and the previous SOTA method, SoMA. 
  }
  \label{fig:teaser}
\vspace{-0.6cm}
\end{figure}

Traditional DGSS approaches focused on data augmentation and domain-invariant feature learning but used backbones trained on limited datasets~\cite{PASTA, GTR, MRFP, SANSAW, RobustNet, BlindNet}, whose knowledge was confined to specific domains and thus had limited generalization capability~\cite{ResNet, MobileNet, ImageNet}. 
With the emergence of Vision Foundation Models (VFMs) such as DINOv2~\cite{DINOv2} and CLIP~\cite{CLIP}—trained on large-scale, diverse datasets that already capture rich and transferable knowledge across domains—the emphasis in DGSS is shifting from diversifying inputs to preserving and efficiently adapting VFM's world knowledge.
To that end, recent studies have explored Parameter-Efficient Fine-Tuning (PEFT) methods for adapting VFMs, achieving competitive performance with minimal computational overhead.
In DGSS, Rein~\cite{Rein} introduced learnable tokens, while SoMA~\cite{SoMA} leveraged Low-Rank Adaptation (LoRA)~\cite{LoRA} to selectively adjust minor components through Singular Value Decomposition (SVD). Meanwhile, Tqdm~\cite{tqdm} and MFuser~\cite{mamba} leveraged Vision-Language Models (VLMs) to enhance cross-domain generalization. 

However, existing methods face limitations in fully exploiting the potential of VFMs. First, while SVD-based approaches such as SoMA~\cite{SoMA} have shown promising results by focusing on minor singular components for preserving pre-trained knowledge, it remains underexplored whether SVD is the most effective decomposition method for adapting Vision Foundation Models. In particular, SVD prioritizes variance preservation, which, while mathematically optimal for data reconstruction, does not necessarily guarantee the most relevant directions for downstream adaptation~\cite{svd_optimal_2, svd_optimal}. Moreover, SoMA adjusts only the minor directions, leaving potentially useful major components untouched during adaptation. This restricted view may limit the model’s capacity to handle complex new tasks or fully exploit the rich representations in VFMs.
Moreover, many LoRA-based methods suffer from limited representational diversity due to learning redundant representations among their basis vectors, which leads to inefficient parameter utilization(e.g., Tab.~\ref{tab:rank_efficiency}, Fig.~\ref{fig:cosinesimilarity}) In domain generalization research, enhancing representational diversity has been shown to improve generalization performance by enabling models to capture a wider range of features~\cite{PEGO, yang2024featurediv, wang2021learningdiv, meng2022attentiondiv}. This issue of representational collapse or rank deficiency in LoRA has been noted and explored in several recent studies~\cite{ReLoRA, HiRA, 3gu2024light, erkim2025improving, kurtz2023group}.

To address these problems, we introduce an initialization strategy based on Rank-Revealing QR Decomposition (RRQR). In contrast to SVD, which finds new orthogonal bases that preserve global variance, 
RRQR selects informative columns directly from the original weight matrix using greedy column pivoting~\cite{rrqr}. At each step, it identifies the column with the largest orthogonal component relative to the previously selected subspace, thereby minimizing redundancy and preserving directional independence.

Since this approach constructs the basis vectors based on columns selected directly from the original weight matrix, the unique structural information held by those columns is well reflected in the new basis. By selecting basis vectors from the actual columns of the weight matrix, RRQR retains localized structural information and preserves the correspondence between weight dimensions and learned representations. This leads to LoRA adapters that are both interpretable and diverse in representation. Importantly, our RRQR-based initialization helps mitigate the representational redundancy often observed among LoRA's basis vectors. RRQR's greedy selection process promotes directional independence, naturally constructing a LoRA adapter with enhanced representational capacity. By recycling structurally informative directions, our method enhances both parameter efficiency and adaptation capacity while preserving the core knowledge embedded in VFMs.
Furthermore, while methods that focus only on minor directions are effective for preserving the VFM's pre-trained knowledge, they can struggle to adapt to new, complex tasks. Recent work such as PiSSA~\cite{Pissa} has demonstrated that effective task adaptation can be achieved by tuning only the major directions of the pretrained weights. Motivated by this, we extend our design with a complementary sub-adapter that carefully refines major directions, further improving generalization performance.

Building upon these insights, we propose \textit{RecycleLoRA}, a novel approach to utilizing pre-trained weights through RRQR decomposition. As demonstrated in Fig.~\ref{fig:teaser}, our main adapter, initialized with minor directions identified by RRQR, achieves state-of-the-art performance by learning diverse and independent features. This demonstrates that strategically recycling these minor directions alone surpasses existing methods. To further enhance performance, we introduce a sub adapter that carefully refines major directions with minimal adjustments, providing complementary improvements. This strategic design enables the two adapters to naturally learn complementary features without additional regularization losses or complex training regimes. As shown in our analysis, the two adapters learn to operate in distinct subspaces and induce different types of modifications in the feature space. 

Experimental results demonstrate that RecycleLoRA achieves top performance in both synthetic-to-real and real-to-real generalization tasks without VLMs or complex architectures. This shows that superior performance in domain generalization can be achieved by effectively exploiting pre-trained subspace structures.

\noindent Our main contributions are as follows.

\begin{itemize}

    \item We propose a novel initialization strategy for LoRA based on Rank-Revealing QR Decomposition, which mitigates representational redundancy by selecting structurally diverse directions from the original weight matrix. This improves both parameter utilization and task-specific adaptability in VFM fine-tuning.

    \item We design a dual-adapter structure that combines a main adapter leveraging minor directions with a sub adapter refining major directions, enabling the model to naturally learn complementary feature representations without explicit regularization.
    
    \item Our method achieves state-of-the-art performance, with 68.95 mIoU in synthetic-to-real generalization and 72.10 mIoU in real-to-real generalization.
    
\end{itemize}

\section{Related Work}

\subsection{Domain Generalized Semantic Segmentation}
Domain Generalized Semantic Segmentation (DGSS) aims to train models that can generalize to unseen target domains without access to target domain data during training. Early approaches primarily focused on alleviating domain shift through data augmentation and adversarial training techniques, but their performance was constrained by the limited representational power of the conventional backbones they relied on, which were often trained on limited datasets~\cite{PASTA, kamann2020increasing, GTR, MRFP, SANSAW, zhong2022adversarial, jia2024dginstyle, niemeijer2024generalization, fahes2024simple, lee2022wildnet, pan2018two, TLDR, wu2022siamdoge}.

The emergence of Vision Foundation Models (VFMs) has introduced new paradigms for DGSS. SoMA~\cite{SoMA} introduces a method that leverages the subspace structure of pre-trained weights by selectively tuning minor singular components through singular value decomposition, effectively preserving the generalization capacity of VFMs while acquiring task-specific knowledge. Rein~\cite{Rein} proposes a parameter-efficient approach utilizing learnable tokens that refine feature maps layer-by-layer, enabling instance-level refinement within the backbone architecture. Meanwhile, methods such as MFuser~\cite{mamba} and tqdm~\cite{tqdm} have improved generalization performance by leveraging the domain-invariant properties of text information based on VLMs. While recent VFM-based DGSS methods have primarily focused on preserving pre-trained knowledge, approaches that systematically recycle and exploit their rich internal subspace structures remain underexplored.

\subsection{Vision Foundation Models}
Vision Foundation Models have emerged as powerful tools for various computer vision tasks, offering strong generalization capabilities across diverse domains. Among prominent VFMs, DINOv2~\cite{DINOv2} utilizes self-supervised learning techniques to learn robust visual representations from diverse visual data, enabling broad applicability across various downstream tasks. EVA02-CLIP~\cite{EVA02} is a Vision-Language Model that provides robust, domain-invariant representations by aligning visual features with textual semantics. CLIP~\cite{CLIP} has established itself as a foundational Vision-Language Model through joint training on image-text pairs, enabling zero-shot classification and cross-modal understanding.

\subsection{Parameter-Efficient Fine-Tuning}
Parameter-Efficient Fine-Tuning (PEFT) has become a standard for adapting large models, as it enables fine-tuning with only a small fraction of the total parameters. Among these techniques, Low-Rank Adaptation (LoRA)~\cite{LoRA} is a prominent method that freezes the original weights and injects trainable, low-rank matrices to model weight updates, achieving comparable performance to full fine-tuning with high parameter efficiency.

Recent developments in PEFT have explored more sophisticated initialization strategies that leverage subspace structures. SoMA~\cite{SoMA} utilizes singular value decomposition to identify and tune minor singular components, specifically targeting the less dominant singular values while preserving the major ones to maintain the pre-trained knowledge. This approach focuses on knowledge preservation by selectively modifying the subspace components that contribute less to the original representation, primarily concentrating on the minor components. Methods like PiSSA~\cite{Pissa} also use singular value decomposition for initialization to leverage the principal directions of weight matrices. These approaches have shown that consideration of the underlying structure in pre-trained weights can significantly impact the effectiveness of low-rank adaptation.

Existing PEFT methods for domain generalization exhibit two main limitations. First, they tend to focus on preserving pre-trained knowledge rather than actively exploiting it. Second, many LoRA-based approaches suffer from inefficient parameter utilization and limited representational capacity, which leads to under-utilized subspace information and parameter redundancy~\cite{LoRASculpt, SoRA}. Our approach addresses both of these challenges by systematically recycling subspace components to simultaneously improve LoRA's parameter efficiency and representational capabilities.

\begin{figure*}[h]
  \centering
  \includegraphics[width=\textwidth]{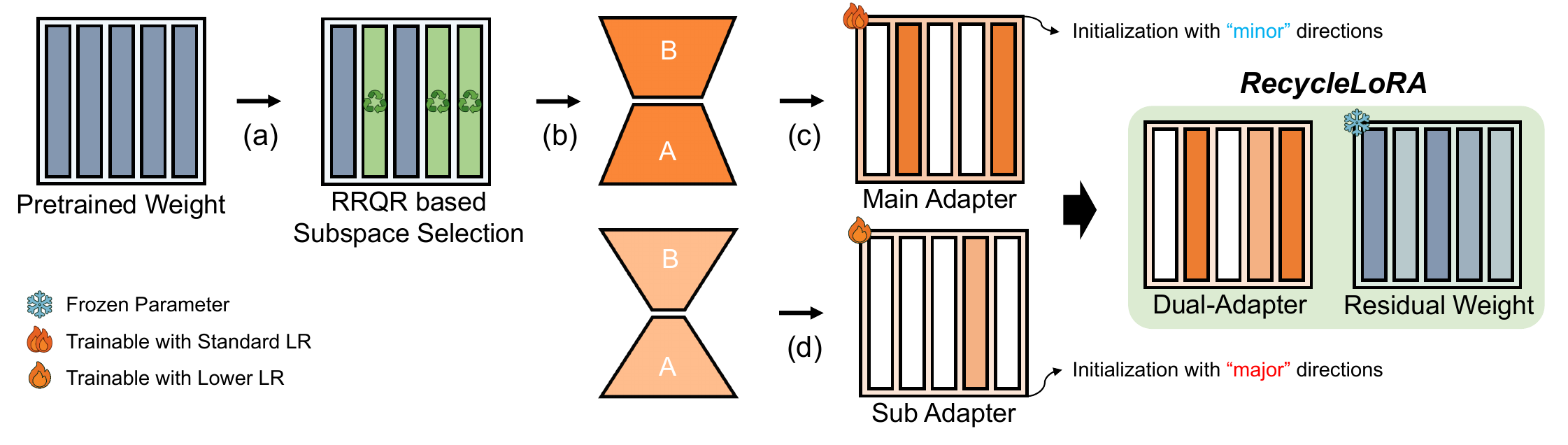}
  \caption{
    \textbf{RecycleLoRA Framework Overview.}
    This figure illustrates the overall workflow of RecycleLoRA. 
    (a) Rank-Revealing QR Decomposition (RRQR) is applied to the pre-trained weight matrix to identify subspace directions ranked by importance. 
    (b) Among the recyclable subspaces selected through RRQR, the minor directions are assigned as initialization values for the main adapter, while the major directions are assigned to the sub adapter. 
    (c) The main adapter's B matrix is initialized with the minor directions, and its A matrix is sparsely initialized by mapping these directions to their corresponding column indices.
    (d) The sub adapter's B matrix is initialized with the major directions, and its A matrix is sparsely initialized by mapping these directions to their corresponding column indices.
  }
  \label{fig:recyclelora_framework}
  \vspace{-0.4cm}
\end{figure*}

\section{Proposed Methods}
In this section, we introduce our proposed method, RecycleLoRA. Section~\ref{preliminaries} provides the necessary technical background on Low-Rank Adaptation (LoRA) and Rank-Revealing QR Decomposition (RRQR), which are foundational to our method. Subsequently, in Section~\ref{RecycleLoRA}, we present the detailed design of RecycleLoRA, validating its effectiveness through an in-depth investigation.
\subsection{Preliminaries}
\label{preliminaries}
\textbf{Low-Rank Adaptation (LoRA).} LoRA is a parameter-efficient fine-tuning technique that models weight updates through trainable low-rank decomposition while freezing pre-trained weights $\mathbf{W}_0 \in \mathbb{R}^{d \times k}$. The weight update $\Delta\mathbf{W}$ is represented as:
\begin{equation}
\mathbf{W} = \mathbf{W}_0 + \Delta\mathbf{W} = \mathbf{W}_0 + \mathbf{BA}
\end{equation}
where $\mathbf{B} \in \mathbb{R}^{d \times r}$, $\mathbf{A} \in \mathbb{R}^{r \times k}$, and the rank $r$ is much smaller than the original dimensions ($r \ll \min(d, k)$), constraining the adaptation to a low-dimensional subspace. The learned low-rank matrices can be merged into the original weights during inference, introducing no additional inference latency.

\noindent \textbf{Rank-Revealing QR Decomposition (RRQR).} The assessment of parameter importance has been a significant research topic across various areas of deep learning. A substantial body of work has established that weight magnitudes serve as effective indicators of parameter importance. Classical pruning methods such as magnitude-based pruning demonstrate that parameters with larger magnitudes typically contribute more significantly to model performance~\cite{learningboth, optimization, 1elesedy2020lottery, 1lee2020layer}. This principle has been extended to various contexts, including structured pruning, where entire channels or layers are ranked by their norm-based importance scores~\cite{2filterspruning, 2he2020learning, 2huang2021rethinking, 2sun2024towards}. Recent advances in parameter-efficient fine-tuning have similarly leveraged magnitude-based importance measures~\cite{LoRASculpt, 3chen2023lorashear, 3gu2024light, 3liu2025lift}. These findings establish weight magnitude as an effective indicator of parameter importance, motivating our adoption of RRQR decomposition, which systematically ranks matrix columns by their norm-based importance. For a matrix $W \in \mathbb{R}^{m \times n}$, the Rank-Revealing QR (RRQR) decomposition is expressed as:
\begin{equation}
\mathbf{WP=QR}
\end{equation}
where $P \in \mathbb{R}^{n \times n}$ is a permutation matrix, $Q \in \mathbb{R}^{m \times n}$ is an orthogonal matrix ($Q^TQ = I$), and $R \in \mathbb{R}^{n \times n}$ is an upper triangular matrix whose diagonal elements capture the magnitude of each column's orthogonal component, and whose off-diagonal elements encode the dependencies between columns.
At each step $k$, the algorithm selects the next column from the set of remaining columns. The chosen column is the one that has the largest norm after being projected onto the orthogonal complement of the subspace spanned by the previously selected columns. Specifically, given the already selected columns ${WP_1, \ldots, WP_{k-1}}$, the algorithm chooses the next column $WP_k$ that maximizes:
\begin{equation}
\left\| \mathbf{W}\mathbf{P}_k - \text{proj}_{\text{span}(\mathbf{W}\mathbf{P}_1, \ldots, \mathbf{W}\mathbf{P}_{k-1})}(\mathbf{W}\mathbf{P}_k) \right\|_2
\end{equation}
where $\text{proj}_{\text{span}(\cdot)}$ denotes orthogonal projection onto the subspace spanned by the argument.
This greedy selection process typically produces a strong tendency for the diagonal elements of $R$ to satisfy:
\begin{equation}
|r_{11}| \geq |r_{22}| \geq \cdots \geq |r_{nn}|
\end{equation}
indicating that most of the matrix energy is concentrated in the leading components.
The permutation matrix $P$ records this importance ordering, where $P[i]$ indicates the original column index of the $i$-th most important direction. The orthogonal matrix $Q$ provides the corresponding orthonormal basis, where each column $q_i$ represents the normalized direction of the orthogonal component of the $P[i]$-th column.
Thus, RRQR provides two key insights: the permutation matrix $P$ identifies the importance ordering of original columns, while the orthogonal matrix $Q$ provides the corresponding geometric directions. This structural characterization enables effective LoRA initialization through sparse mapping of specific input dimensions, enhancing representation diversity and parameter utilization.

\subsection{RecycleLoRA}
\label{RecycleLoRA} RecycleLoRA is a dual-adapter methodology that uses RRQR decomposition to separate pre-trained weights into minor and major directions, which initialize a main and sub adapter, respectively (Figure~\ref{fig:recyclelora_framework}). To preserve the initial output of the pre-trained weights, we construct a residual matrix by subtracting the initial adapter values from the original weights before training. This matrix is then frozen, so that only the two adapters are trained. To demonstrate the effectiveness of our approach, we compare our method against SoMA~\cite{SoMA}, which is the previous state-of-the-art and a LoRA-based method.

\begin{table}[t]
\centering
\colorlet{LightGray}{gray!40}
\caption{
Comparison of $\ell_2$-norm statistics between selected and non-selected columns in LoRA matrix A after training.}
\label{tab:lora_norm_summary}
\resizebox{1.0\columnwidth}{!}{
{\begin{tabular}{l|c|c|c|c}
\toprule
\multicolumn{5}{c}{\textmd{\textbf{LoRA Column‑wise Norm Statistics}}} \\ \midrule
 & \textmd{Mean$\,\ell_2$ norm} & \textmd{Mean$\,\ell_2$ norm} & \textmd{Average} & \textmd{Maximum} \\
 & \textmd{(selected)} & \textmd{(non‑selected)} & \textmd{ratio} & \textmd{ratio} \\ 
\midrule
All layers & 0.138245 & 0.113566 & 1.22\,$\times$ & 1.63\,$\times$ \\
\bottomrule
\end{tabular}}
}
\end{table}

\begin{table}[t]
\centering
\colorlet{LightGray}{gray!40}
\caption{Effective rank and rank efficiency comparison. Rank efficiency is calculated as the ratio of effective rank to target rank, measuring parameter utilization. RecycleLoRA consistently achieves higher efficiency than SoMA across different rank settings.}
\label{tab:rank_efficiency}
\resizebox{1.0\columnwidth}{!}{
{\begin{tabular}{c|c|c|c|c}
\toprule
\multicolumn{5}{c}{\textmd{\textbf{Effective Rank \& Rank‑Efficiency Statistics}}}\\ \midrule
\multirow{2}{*}{\textmd{Target rank}} & \multicolumn{2}{c|}{\textmd{RecycleLoRA (Ours)}} & \multicolumn{2}{c}{\textmd{SoMA}} \\
 & \textmd{Effective rank} & \textmd{Rank efficiency} & \textmd{Effective rank} & \textmd{Rank efficiency} \\ \midrule
16 & 13.60 & 0.850 & 9.78 & 0.611 \\ \midrule
32 & 24.65 & 0.770 & 20.80 & 0.650 \\
\bottomrule
\end{tabular}}
}
\end{table}
\noindent \textbf{Main Adapter.} RecycleLoRA leverages RRQR decomposition to design a novel initialization strategy that systematically recycles pre-trained knowledge from VFMs while simultaneously enhancing LoRA's representational efficiency for improved domain generalization. Specifically, we perform RRQR decomposition on each linear layer's weight matrix $\mathbf{W}_0 \in \mathbb{R}^{d \times k}$ to obtain:
\begin{equation}
\mathbf{W}_0\mathbf{P} = \mathbf{Q}\mathbf{R}
\end{equation}
where $\mathbf{Q} \in \mathbb{R}^{d \times k}$ is an orthogonal matrix and $\mathbf{R} \in \mathbb{R}^{k \times k}$ is upper triangular. In practice, the permutation matrix $\mathbf{P}$ is returned as an index array $P \in \mathbb{N}^k$, where $P[i]$ indicates the original column index of the $i$-th most important direction.
\noindent The main adapter is initialized as:
\begin{equation}
\mathbf{B}_{\text{main}} = \mathbf{Q}[:, -r_{\text{main}}:]
\end{equation}
\begin{equation}
\mathbf{A}_{\text{main}}[i, j] = \begin{cases}
1, & \text{if } j = P[k-r_{\text{main}}+i] \\
0, & \text{otherwise}
\end{cases}
\end{equation}
where $i \in \{0, 1, ..., r_{\text{main}}-1\}$ and $j \in \{0, 1, ..., k-1\}$. This sparse initialization is designed to focus initial changes on specific input dimensions identified by RRQR.

To validate the effectiveness of this initialization strategy, we analyzed the evolution of LoRA parameters throughout training. Table~\ref{tab:lora_norm_summary} presents the column-wise norms of the $\mathbf{A}$ matrix after training, revealing that columns containing sparsely initialized positions with value 1 maintain norms that are on average $1.22\times$ larger, and up to $1.63\times$ larger, than columns initialized entirely to 0. This suggests that parameter updates during training were relatively concentrated on the initially selected directions, demonstrating that our sparse initialization maintains structural bias to some extent throughout the training process.

\begin{figure}[t]
  \centering
  \includegraphics[width=\columnwidth]{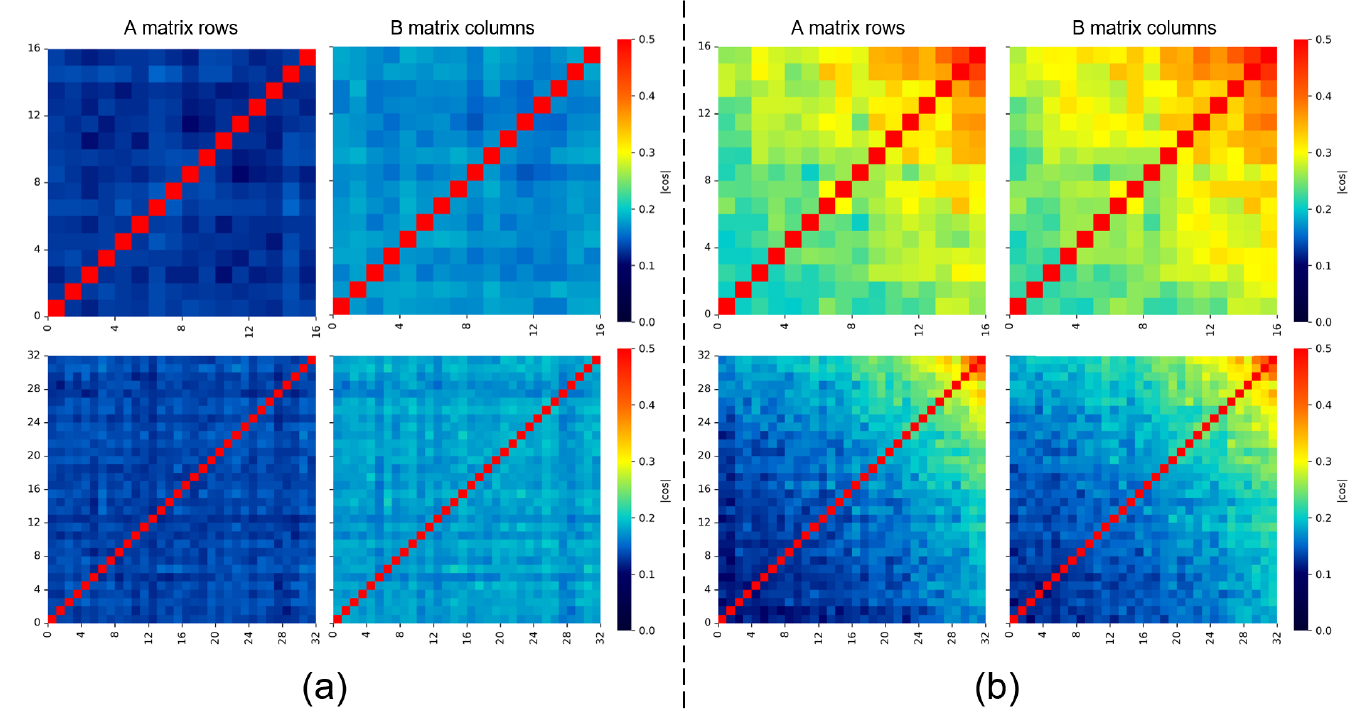}
  \caption{
    Cosine similarity heatmaps of LoRA components for (a) RecycleLoRA and (b) SoMA at different ranks (r=16, 32). 
    Left: pairwise similarity among rows of $\mathbf{A}$. Right: pairwise similarity among columns of $\mathbf{B}$.
    Darker blue colors represent lower similarity.
  }
  \label{fig:cosinesimilarity}
  \vspace{-0.6cm}
\end{figure}

Furthermore, we evaluated the diversity of learned representations and the efficiency of parameter utilization. Figure~\ref{fig:cosinesimilarity} visualizes the cosine similarity between LoRA components for both SoMA and RecycleLoRA, showing that RecycleLoRA consistently exhibits lower similarity across various rank settings. Specifically, we focused our analysis on the dimensions that actually determine the rank in LoRA's low-rank structure: the similarity between rows of $\mathbf{A} \in \mathbb{R}^{r \times k}$ and the similarity between columns of $\mathbf{B} \in \mathbb{R}^{d \times r}$. This is because in LoRA's weight update $\Delta\mathbf{W} = \mathbf{BA}$, the $i$-th row of $\mathbf{A}$ and the $i$-th column of $\mathbf{B}$ together form a single low-rank component. Therefore, low similarity among rows of $\mathbf{A}$ and low similarity among columns of $\mathbf{B}$ indicate that each low-rank component captures distinct, independent features. This reduced similarity encourages each low-rank component to learn more independent and distinctive features, thereby improving the utilization efficiency of limited parameters.

This diverse representation learning directly translates to enhanced model expressiveness. As shown in Table~\ref{tab:rank_efficiency}, RecycleLoRA consistently achieves higher Effective Rank~\cite{effectiverank} compared to SoMA. The Effective Rank quantifies the dimensional richness of learned representations, and recent studies have shown that higher Effective Rank correlates with improved representational capacity and generalization performance~\cite{HiRA, ReLoRA, erkim2025improving, erffeng2022rank}. The consistent improvements observed across various settings demonstrate that our method utilizes LoRA's limited parameters more efficiently to enhance representational capacity.

\begin{figure}[t!]
  \centering
  \includegraphics[width=0.9\columnwidth]{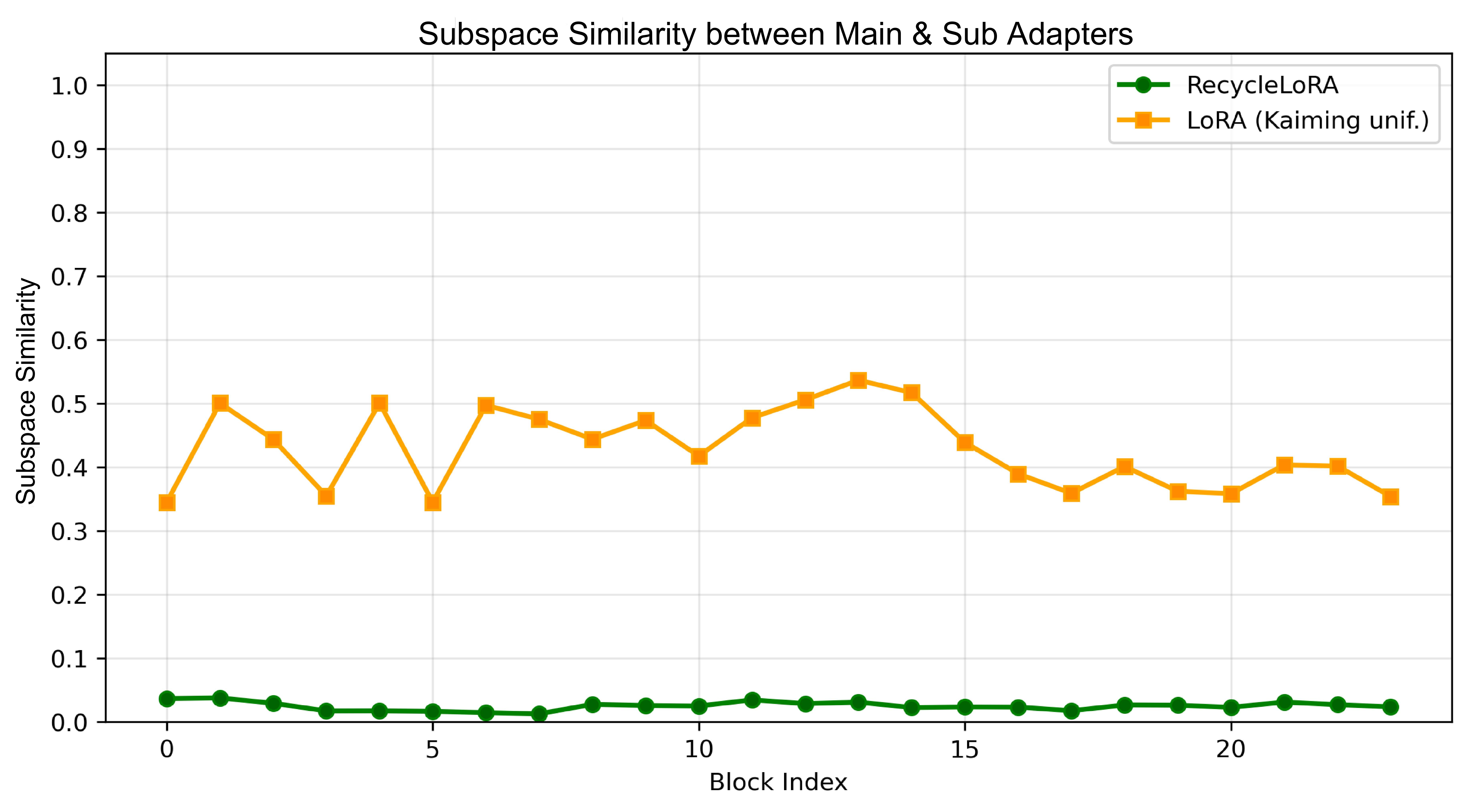}
  \caption{
    Block-wise subspace similarity $\phi$ between main and sub adapters measured using Grassmann Distance on the low-rank matrices. 
    Lower values indicate more orthogonal subspaces.
  }
  \label{fig:subspace_similarity}
  \vspace{-0.4cm}
\end{figure}

\noindent \textbf{Sub Adapter.} A key insight in the design of RecycleLoRA is that different subspace components within VFM weights can contribute complementarily to domain generalization performance. Building on this observation, we introduce a sub adapter that complements the main adapter.

By investigating recent LoRA initialization strategies, we uncovered a critical yet underexplored relationship between the choice of initialization subspace and the optimal learning rate. Specifically, we observed that methods that initialize LoRA with major directions, such as PiSSA~\cite{Pissa}, tend to adopt lower learning rates. We can infer that this is because major directions encode the Vision Foundation Model’s core, generalizable knowledge, making them sensitive to large updates that could risk catastrophic forgetting. A lower learning rate thus enables a careful refinement of these critical components, preserving foundational knowledge while adapting to the new task. Conversely, we observed that approaches using minor directions, like SoMA~\cite{SoMA}, tend to employ relatively higher learning rates. Since these minor directions contribute less to the model’s pretrained capabilities, they can provide a safer subspace for learning new representations. A higher learning rate allows for more aggressive and efficient adaptation within this subspace without jeopardizing the model's core representations. This suggests an inherent relationship between the initialization strategy and the optimal learning rate, rooted in the trade-off between knowledge preservation and task adaptation.

Motivated by this observation, we investigated the interplay between initialization methods and learning rates. Our experiments, detailed in Section~\ref{ablation} (Tab ~\ref{tab: lr}), suggest a tendency where the optimal learning rate is contingent upon the nature of the initialized directions. Specifically, the sub adapter, initialized with RRQR's top directions, tended to exhibit improved performance at a lower learning rate (5e-5), which implies that the major directions encoding the VFM's core knowledge require more careful optimization. In contrast, both the main adapter initialized with minor directions and standard LoRA with Kaiming initialization achieved peak performance at the standard learning rate (1e-4), with their performance declining when the learning rate was reduced. Notably, this performance drop was more pronounced for the main adapter than for the Kaiming-initialized LoRA. This result suggests that the minor directions provide a safer subspace for learning new, task-specific features, thereby benefiting from more aggressive updates. These contrasting findings validate our design choice of employing a differentiated learning rate scheme in our dual-adapter architecture, where each adapter is optimized according to the sensitivity of its assigned subspace.

\begin{figure}[t!]
  \centering
  \includegraphics[width=\columnwidth]{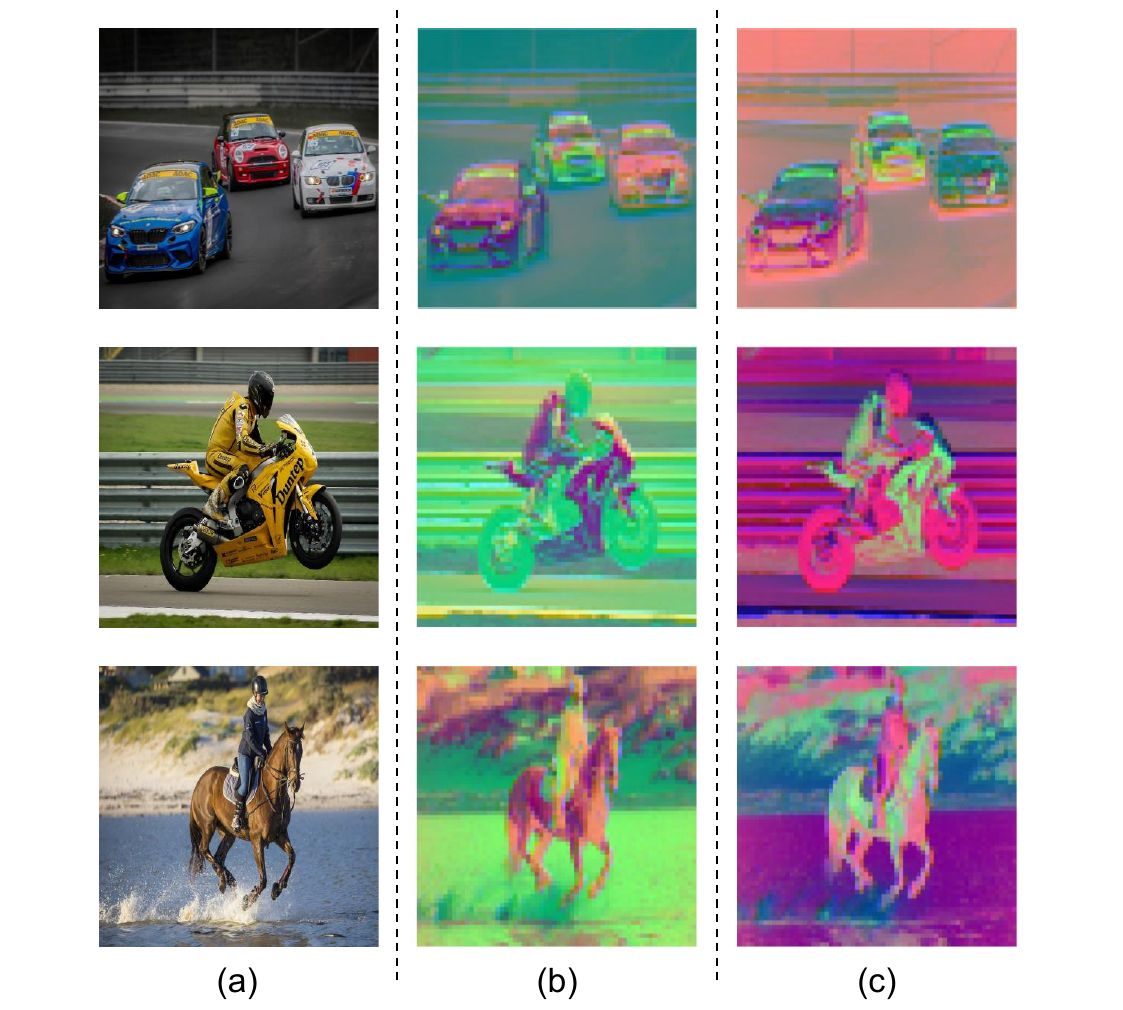}
  \caption{
    Visualization of adapter-induced feature modifications via PCA projection. (a) Input. (b) main adapter PCA Visualization. (c) sub adapter PCA Visualization. The divergent activation patterns reveal complementary feature learning between the dual adapters.
  }
  \label{fig:pca_visualization}
  \vspace{-0.4cm}
\end{figure}

Based on these findings, we incorporate a sub adapter that leverages the top directions from RRQR. While our main adapter alone already surpasses existing state-of-the-art methods (as will be demonstrated in Table~\ref{tab:adapter_ablation}, Section 4.3), the sub adapter provides complementary performance improvements.

\noindent The sub adapter is initialized as:
\begin{equation}
\mathbf{B}_{\text{sub}} = \mathbf{Q}[:, :r_{\text{sub}}]
\end{equation}

\begin{equation}
\mathbf{A}_{\text{sub}}[i, j] = \begin{cases}
1, & \text{if } j = P[i] \\
0, & \text{otherwise}
\end{cases}
\end{equation}
The sub adapter is trained with a much smaller rank (32$\rightarrow$4) and lower learning rate (1e-4$\rightarrow$5e-5) than the main adapter, allowing careful adjustment of the major directions in the pre-trained weights.

To verify that the two adapters learn distinct representations, we conducted several analyses. Figure~\ref{fig:subspace_similarity} presents the subspace similarity measured using Grassmann Distance. Specifically, we compute the similarity between the subspaces spanned by the left singular vectors ($\mathbf{U}_{\text{main}}, \mathbf{U}_{\text{sub}}$) from the SVD of each adapter's matrix $\mathbf{A}$:
\begin{equation}
\phi = \frac{\|\mathbf{U}_{\text{main}}^T\mathbf{U}_{\text{sub}}\|_F^2}{\min(r_{\text{main}}, r_{\text{sub}})} \in [0, 1]
\end{equation}
This metric, introduced in the LoRA paper~\cite{LoRA}, ranges from 0 to 1, where 1 indicates identical subspaces and 0 signifies complete orthogonality. As shown in Figure~\ref{fig:subspace_similarity}, our RecycleLoRA framework yields a consistently low similarity between the main and sub adapters. As a baseline, we trained a dual-adapter model that, while initialized with the standard Kaiming method, otherwise shared the identical configuration of RecycleLoRA, including the differentiated rank and learning rate settings for the main and sub adapters. This baseline exhibited significantly higher subspace similarity than RecycleLoRA. This comparative analysis demonstrates that our RRQR-based initialization is instrumental in guiding the adapters to operate in distinct, nearly orthogonal subspaces.

Furthermore, Figure~\ref{fig:pca_visualization} presents PCA projections of the feature differences produced by the main and sub adapters in the final block of DINOv2. The main adapter induces localized, salient modifications around foreground objects, whereas the sub adapter yields broader shifts that spread across the background. These complementary patterns support the claim that the RRQR-based initialization and the differentiated rank and learning-rate design steer the two adapters toward learning complementary  representations.

This integration of RRQR-based initialization, differentiated rank allocation, and carefully tuned learning rates enables RecycleLoRA to systematically exploit VFM's multi-domain knowledge, achieving superior domain generalization performance without additional regularization.

\vspace{-0.2cm}
\section{Experiments}
\vspace{-0.1cm}
\subsection{Experimental Settings}
\vspace{-0.1cm}
\textbf{Datasets.} We evaluate the effectiveness of RecycleLoRA using widely adopted benchmark datasets for Domain Generalized Semantic Segmentation. For synthetic data, we use GTAV~\cite{GTAV}, which consists of 12,403 training images, 6,382 validation images, and 6,181 test images. For real-world data, we employ Cityscapes~\cite{Cityscapes} with 2,975 training images and 500 validation images, Berkeley Deep Driving dataset~\cite{BDD} with 1,000 validation images, and Mapillary~\cite{Mapillary} with 2,000 validation images.

\noindent \textbf{Implementation Details.} We use DINOv2-Large as the backbone and Mask2Former as the segmentation head. RecycleLoRA is applied to all linear layers within the self-attention modules and MLP layers of the transformer. The main adapter uses a rank of 32, while the sub adapter's rank is set to 4 for the synthetic-to-real setting and 2 for the real-to-real setting. The learning rate multipliers for the main and sub adapters are 1.0 and 0.5, respectively. All experiments use 512×512 cropped images for training.
\begin{table}[t]
\centering
\colorlet{LightGray}{gray!30}
\caption{Domain generalization results (mIoU \%) under the synthetic-to-real setting. Our method is highlighted in \colorbox[HTML]{e0e0e0}{gray}. \textbf{Bold} and \underline{underlined} indicate best and second-best results.}
\label{tab: g2cbm}
\resizebox{1\columnwidth}{!}{
{\begin{tabular}{l|c|c|c|c|c|c}
\toprule
\multicolumn{7}{c}{\textmd{\textbf{Synthetic-to-Real Generalization}}} \\ \midrule
\multirow{2}{4em}{Method} & \multirow{2}{*}{Venue} & \multirow{2}{*}{Backbone} &
\multicolumn{4}{c}{\textmd{\textit{Trained on GTAV}}} \\
 & & & $\rightarrow$Citys. & $\rightarrow$BDD & $\rightarrow$Map. & Avg. \\ 
\midrule
CLOUDS~\cite{clouds} & CVPR2024 & CLIP-CN-L & 60.20 & 57.40 & 67.00 & 61.50 \\
VLTSeg~\cite{vltseg} & ACCV2024 & EVA02-L & 65.30 & 58.30 & 66.00 & 63.20 \\
DoRA~\cite{dora} & ICML2024 & DINOv2-L & 66.12 & 59.31 & 67.07 & 64.17 \\
VPT~\cite{VPT} & ECCV2022 & DINOv2-L & 68.75 & 58.64 & 68.32 & 65.24 \\
SET~\cite{SET} & TIP2021 & DINOv2-L & 68.06 & 61.64 & 67.68 & 65.79 \\
tqdm~\cite{tqdm} & ECCV2024 & EVA02-L & 68.88 & 59.18 & 70.10 & 66.05 \\
Rein$^{\dagger}$~\cite{Rein} & CVPR2024 & DINOv2-L & 69.19 & 60.01 & 69.06 & 66.09 \\
FADA~\cite{FADA} & NeurIPS2024 & DINOv2-L & 68.23 & 61.94 & 68.09 & 66.09 \\
AdaptFormer~\cite{AdaptFormer} & NeurIPS2022 & DINOv2-L & 70.10 & 59.81 & 68.77 & 66.23 \\
PEGO~\cite{PEGO} & ECCV2024 & DINOv2-L & 68.86 & 61.44 & 68.61 & 66.30 \\
SSF~\cite{ssf} & NeurIPS2022 & DINOv2-L & 68.97 & 61.30 & 68.77 & 66.35 \\
LoRA~\cite{LoRA} & ICLR2022 & DINOv2-L & 70.13 & 60.13 & 70.42 & 66.89 \\
DepthForge~\cite{depthforge} & ICCV2025 & DINOv2-L & 69.04 & \underline{62.82} & 69.22 & 67.03 \\
DPMFormer~\cite{dpmformer} & ICCV2025 & EVA02-L & 70.08 & 60.48 & 70.66 & 67.07 \\
Mfuser~\cite{mamba} & CVPR2025 & EVA02-L & 70.19 & \textbf{63.13} & 71.28 & 68.20 \\
SoMA~\cite{SoMA} & CVPR2025 & DINOv2-L & \underline{71.82} & 61.31 & \underline{71.67} & \underline{68.27} \\
\rowcolor[HTML]{e0e0e0} RecycleLoRA & - & DINOv2-L & \textbf{73.01} & 61.77 & \textbf{72.07} & \textbf{68.95} \\
\midrule
\multirow{2}{4em}{Method} & \multirow{2}{*}{Venue} & \multirow{2}{*}{Backbone} &
\multicolumn{4}{c}{\textmd{\textit{Trained on GTAV + Synthia}}} \\
 & & & $\rightarrow$Citys. & $\rightarrow$BDD & $\rightarrow$Map. & Avg. \\ 
\midrule
Rein$^{\dagger}$~\cite{Rein} & CVPR2024 & DINOv2-L & 72.17 & 61.53 & 70.69 & 68.13 \\
SoMA~\cite{SoMA} & CVPR2025 & DINOv2-L & \underline{73.16} & \textbf{61.90} & \textbf{72.73} & \underline{69.26} \\
\rowcolor[HTML]{e0e0e0} RecycleLoRA & - & DINOv2-L & \textbf{73.71} & \underline{61.87} & \underline{72.68} & \textbf{69.42} \\
\midrule
\multirow{2}{4em}{Method} & \multirow{2}{*}{Venue} & \multirow{2}{*}{Backbone} &
\multicolumn{4}{c}{\textmd{\textit{Trained on GTAV + Synthia + UrbanSyn}}} \\
 & & & $\rightarrow$Citys. & $\rightarrow$BDD & $\rightarrow$Map. & Avg. \\ 
\midrule
Full Fine-Tuning & - & DINOv2-L & 75.90 & 60.93 & 72.80 & 69.88 \\
SoMA~\cite{SoMA} & CVPR2025 & DINOv2-L & \underline{77.33} & \underline{62.78} & \textbf{74.93} & \underline{71.68} \\
\rowcolor[HTML]{e0e0e0} RecycleLoRA & - & DINOv2-L & \textbf{78.66} & \textbf{63.46} & \underline{74.83} & \textbf{72.32} \\
\bottomrule
\end{tabular}}
}
\vspace{-0.4cm}
\end{table}
\subsection{Comparison with State-of-the-Art Methods}
To demonstrate the effectiveness of our method, we compare RecycleLoRA against existing state-of-the-art DGSS methods. For a fair comparison, some methods are reimplemented using publicly available official checkpoints, and these reimplemented results are denoted with ${\dagger}$.

\noindent \textbf{Synthetic-to-Real Generalization.} We compare RecycleLoRA against a wide range of recent state-of-the-art methods. As shown in Table~\ref{tab: g2cbm}, in the setting where models are trained on synthetic GTAV data and evaluated on real-world Cityscapes, BDD, and Mapillary data, RecycleLoRA achieves state-of-the-art performance, outperforming all existing methods. Notably, our method achieves 73.01 mIoU on GTAV$\rightarrow$Cityscapes, representing a substantial improvement of 1.19 mIoU over the previous best-performing method. It also achieves a 0.4 mIoU improvement on GTAV$\rightarrow$Mapillary, establishing state-of-the-art performance with an average improvement of 0.68 mIoU. To further assess the robustness and scalability of our method, we also evaluate RecycleLoRA in a multi-source generalization setting where the model is trained on combined synthetic datasets. As detailed in Table~\ref{tab: g2cbm}, when trained on GTAV and Synthia, RecycleLoRA achieves an average mIoU of 69.42, surpassing the previous state-of-the-art method, SoMA~\cite{SoMA}. We further extend the experiment by adding UrbanSyn to the training sources. In this three-source setting, RecycleLoRA again demonstrates superior performance, achieving an average mIoU of 72.32 and widening its performance gap over SoMA. These results confirm that RecycleLoRA effectively leverages multiple source domains and maintains its strong generalization capabilities as the diversity of training data increases.

\noindent \textbf{Real-to-Real Generalization.} In this setting, our method is compared with various competitive approaches. Table~\ref{tab: c2bm} shows that in the real-to-real scenario, where models are trained on Cityscapes and evaluated on BDD and Mapillary, RecycleLoRA consistently demonstrates superior performance, achieving state-of-the-art performance with an average improvement of 0.23 mIoU. These results confirm that our method effectively handles not only synthetic-to-real domain gaps but also variations between different real-world domains.
\begin{table}[t]
\centering
\colorlet{LightGray}{gray!15}
\caption{Domain generalization results (mIoU \%) under the real-to-real setting. Our method is highlighted in \colorbox[HTML]{e0e0e0}{gray}. \textbf{Bold} and \underline{underlined} indicate best and second-best results.}
\label{tab: c2bm}
\resizebox{0.9\columnwidth}{!}{
{\begin{tabular}{l|c|c|c|c|c}
\toprule
\multicolumn{6}{c}{\textmd{\textbf{Real-to-Real Generalization}}} \\ \midrule
\multirow{2}{4em}{Method} & \multirow{2}{*}{Venue} & \multirow{2}{*}{Backbone} &
\multicolumn{3}{c}{\textmd{\textit{Trained on Cityscapes}}} \\
 & & & $\rightarrow$BDD & $\rightarrow$Map. & Avg. \\ 
\midrule
HGFormer~\cite{hgformer} & CVPR2023 & Swin-L & 61.50 & 72.10 & 66.80 \\
CMFormer~\cite{cmformer} & AAAI2024 & Swin-L & 62.60 & 73.60 & 68.10 \\
PDAF~\cite{pdaf} & ICCV2025 & Swin-L & 63.00 & 74.10 & 68.55 \\
SET~\cite{SET} & TIP2021 & DINOv2-L & 65.07 & 75.67 & 70.37 \\
VLTSeg~\cite{vltseg} & ACCV2024 & EVA02-L & 64.40 & 76.40 & 70.40 \\
tqdm~\cite{tqdm} & ECCV2024 & EVA02-L & 64.72 & 76.15 & 70.44 \\
DPMFormer~\cite{dpmformer} & ICCV2025 & EVA02-L & 64.20 & 76.67 & 70.44 \\
FADA~\cite{FADA} & NeurIPS2024 & DINOv2-L & 65.12 & 75.86 & 70.49 \\
Rein$^{\dagger}$~\cite{Rein} & CVPR2024 & DINOv2-L & 66.53 & 75.18 & 70.86 \\
DepthForge~\cite{depthforge} & ICCV2025 & DINOv2-L & 66.19 & 75.93 & 71.06 \\
SoMA~\cite{SoMA} & CVPR2025 & DINOv2-L & \textbf{67.02} & 76.45 & 71.74 \\
MFuser~\cite{mamba} & CVPR2025 & EVA02-L & 65.81 & \textbf{77.93} & \underline{71.87} \\
\rowcolor[HTML]{e0e0e0} RecycleLoRA & - & DINOv2-L & \underline{66.65} & \underline{77.54} & \textbf{72.10} \\
\bottomrule
\end{tabular}}
}
\vspace{-0.1cm}
\end{table}
\begin{table}[t]
\centering
\colorlet{LightGray}{gray!15}
\caption{Ablation study on the Main and Sub Adapter components of RecycleLoRA. Best results in \textbf{bold}.}
\label{tab:adapter_ablation}
\resizebox{1.0\columnwidth}{!}{
{\begin{tabular}{cc|ccccc}
\toprule
{Main} &
{Sub} &
{Params.} &
$\rightarrow$Citys. &
$\rightarrow$BDD &
$\rightarrow$Map. &
Avg. \\
\midrule
\raisebox{-0.2em} &
\raisebox{-0.2em}{\CheckmarkBold} &
1.6M & 70.64 & 60.56 & 71.11 & 67.44\\
\raisebox{-0.2em}{\CheckmarkBold} &
\raisebox{-0.2em} &
12.6M & 72.92 & 61.22 & 71.75 & 68.63\\
\rowcolor[HTML]{e0e0e0}
\raisebox{-0.2em}{\CheckmarkBold} &
\raisebox{-0.2em}{\CheckmarkBold} &
14.2M & \textbf{73.01} \scriptsize \textcolor{red}{($\uparrow$ 0.09)} & \textbf{61.77} \scriptsize \textcolor{red}{($\uparrow$ 0.55)} & \textbf{72.07} \scriptsize \textcolor{red}{($\uparrow$ 0.32)} & \textbf{68.95} \scriptsize \textcolor{red}{($\uparrow$ 0.32)}\\
\bottomrule
\end{tabular}}
}
\vspace{-0.4cm}
\end{table}
\subsection{Ablation Studies}
\label{ablation}
To validate our design choices, we conduct a series of ablation studies in the synthetic-to-real generalization setting.

\noindent \textbf{Components Analysis.} Table~\ref{tab:adapter_ablation} presents an analysis of each component's contribution. Remarkably, using only the main adapter achieves an average mIoU of 68.63, which already surpasses all previous state-of-the-art methods, including SoMA. This result highlights that our RRQR-based initialization strategy for the main adapter is highly effective on its own. The addition of the sub adapter further boosts performance across all target domains, reaching an average of 68.95 mIoU. This confirms that the two adapters work in a complementary manner to maximize domain generalization performance.

\begin{table}[t]
\centering
\colorlet{LightGray}{gray!15}
\caption{Comparative analysis of learning rate sensitivity across initialization strategies. Standard LoRA employs Kaiming initialization while RecycleLoRA's Sub Adapter utilizes RRQR top-ranked directions. The best result for each method is shown in \textbf{bold}.}
\label{tab: lr}
\resizebox{1\columnwidth}{!}{
{\begin{tabular}{l|c|c|c|c|c}
\toprule
Method & lr. & $\rightarrow$Citys. & $\rightarrow$BDD & $\rightarrow$Map. & Avg. \\ 
\midrule
Main Adapter & 1e-4 & \textbf{72.92} & 61.22 & \textbf{71.75} & \textbf{68.63} \\
\rowcolor[HTML]{e0e0e0} Main Adapter & 5e-5 & 69.46 \scriptsize \textcolor{blue}{($\downarrow$ 3.46)} & \textbf{62.22} \scriptsize \textcolor{red}{($\uparrow$ 1.00)} & 68.23 \scriptsize \textcolor{blue}{($\downarrow$ 3.52)} & 66.64 \scriptsize \textcolor{blue}{($\downarrow$ 1.99)} \\
\midrule
LoRA & 1e-4 & \textbf{70.36} & \textbf{60.21} & \textbf{69.95} & \textbf{66.84} \\
\rowcolor[HTML]{e0e0e0} LoRA & 5e-5 & 69.40 \scriptsize \textcolor{blue}{($\downarrow$ 0.96)} & 59.62 \scriptsize \textcolor{blue}{($\downarrow$ 0.59)} & 69.20 \scriptsize \textcolor{blue}{($\downarrow$ 0.75)} & 66.07 \scriptsize \textcolor{blue}{($\downarrow$ 0.77)} \\
\midrule
Sub Adapter & 1e-4 & 68.60 & \textbf{60.74} & 67.54 & 65.63 \\
\rowcolor[HTML]{e0e0e0} Sub Adapter & 5e-5 & \textbf{70.64} \scriptsize \textcolor{red}{($\uparrow$ 2.04)} & 60.56 \scriptsize \textcolor{blue}{($\downarrow$ 0.18)} & \textbf{71.11} \scriptsize \textcolor{red}{($\uparrow$ 3.57)} & \textbf{67.44} \scriptsize \textcolor{red}{($\uparrow$ 1.81)} \\
\bottomrule
\end{tabular}}
}
\vspace{-0.2cm}
\end{table}

\begin{table}[t]
    \centering
    \centering
\colorlet{LightGray}{gray!15}
\caption{Domain generalization performance (mIoU, \%) on the EVA02-L backbone under the synthetic-to-real setting. \textbf{Bold} and \underline{underlined} indicate the best and second-best results, respectively.}
\label{tab: backbone_ablation}
\resizebox{1\columnwidth}{!}{
{\begin{tabular}{l|c|c|c|c|c|c}
\toprule
\multicolumn{7}{c}{\textmd{\textbf{Synthetic-to-Real Generalization}}} \\ \midrule
\multirow{2}{4em}{Method} & \multirow{2}{*}{Venue} & \multirow{2}{*}{Backbone} &
\multicolumn{4}{c}{\textmd{\textit{Trained on GTAV}}} \\
 & & & $\rightarrow$Citys. & $\rightarrow$BDD & $\rightarrow$Map. & Avg. \\ 
\midrule
Rein~\cite{Rein} & CVPR2024 & EVA02-L & 65.30 & 60.50 & 64.90 & 63.60 \\
FADA~\cite{FADA} & NeurIPS2024 & EVA02-L & 66.70 & \textbf{61.90} & 66.10 & 64.90 \\
DepthForge~\cite{depthforge} & ICCV2025 & EVA02-L & 68.00 & \underline{61.70} & 67.50 & \underline{65.73} \\
SoMA~\cite{SoMA} & CVPR2025 & EVA02-L & \underline{68.05} & 60.81 & \underline{68.33} & \underline{65.73} \\
\rowcolor[HTML]{e0e0e0} RecycleLoRA & - & EVA02-L & \textbf{68.95} & 61.37 & \textbf{68.73} & \textbf{66.35} \\
\bottomrule
\end{tabular}}
}

    \vspace{-0.4cm}
\end{table}

\noindent \textbf{Learning Rate Analysis.} To further investigate our hypothesis on the relationship between the initialization strategy and learning rates, we present the analysis in Table ~\ref{tab: lr}. The results indicate that both the main adapter, initialized with RRQR’s minor directions, and the standard LoRA with Kaiming initialization tend to show performance degradation as the learning rate is reduced, with this trend being more pronounced for the main adapter. Interestingly, the sub adapter, which is initialized with RRQR's top directions, exhibits a contrasting and notable pattern of improved performance at a lower learning rate. These observations suggest a potential link between the directions used for initialization and the optimal learning rate, a finding that supports the rationale for applying different learning rates to our main and sub adapters.

\noindent \textbf{Performance on Different VFM backbones.} \noindent To demonstrate its generality, we also evaluated RecycleLoRA on the EVA02-L backbone. As shown in Table~\ref{tab: backbone_ablation}, RecycleLoRA achieves state-of-the-art performance against pure Vision Foundation Model adaptation methods, with VLM-based approaches that leverage textual information being excluded from the comparison~\cite{mamba, tqdm, dpmformer}. This result confirms that our RRQR-based strategy is robust and effective across different VFM architectures.

\vspace{-0.1cm}
\section{Conclusion}
\vspace{-0.1cm}
In this paper, we introduce RecycleLoRA, an approach designed to actively recycle the internal subspace structures of Vision Foundation Models through Rank-Revealing QR decomposition. This strategy enables complementary feature learning by leveraging both minor and major directions, achieving state-of-the-art performance on both synthetic-to-real and real-to-real generalization tasks.

\section{Acknowledgments}
\label{sec:acknowledgement}
This work was supported by the National Research Foundation (NRF) grant funded by the Korea government (MSIT) [RS-2025-00562400] and [RS-2022-NR068754].
% \appendix
% % \section*{Appendix}
% \input{sec/7_supplemental}

{
\small
\bibliographystyle{ieeenat_fullname}
\bibliography{main}
}

% \clearpage
% \setcounter{page}{1}
\maketitlesupplementary

\noindent This supplement provides additional materials omitted from the main text to facilitate a deeper understanding of our proposed RecycleLoRA.

\hypersetup{linkbordercolor=black,linkcolor=black}

\renewcommand{\thesection}{\Alph{section}}
\setcounter{section}{0}

\setlength{\cftbeforesecskip}{0.5em}
\cftsetindents{section}{0em}{1.8em}
\cftsetindents{subsection}{1em}{2.5em}

\etoctoccontentsline{part}{Appendix}
\localtableofcontents
\hypersetup{linkbordercolor=blue,linkcolor=blue}

\section{Implementation Details}
Our method is implemented based on the MMSegmentation codebase. We use DINOv2-Large as the backbone and Mask2Former as the decode head. Following the experimental setups of Rein~\cite{Rein} and SoMA~\cite{SoMA}, we only utilize the default data augmentation provided in Mask2Former~\cite{Mask2Former} to ensure a fair comparison. All models are trained on NVIDIA A6000 GPUs. Further details on hyperparameters are provided in Table~\ref{tab:hyperparameters}. Unless stated otherwise, all experiments in the main paper were conducted using these settings.
\begin{table}[H]
    \centering
    \centering
\resizebox{0.7\columnwidth}{!}{
\begin{tabular}{lcc}
\toprule
\textbf{Hyperparameter} & \textbf{Synthetic-to-Real} & \textbf{Real-to-Real} \\
\midrule
backbone & DINOv2-L & DINOv2-L \\
main rank & 32 & 32 \\
sub rank & 4 & 2 \\
main lr mult. & 1.0 & 1.0 \\
sub lr mult. & 0.5 & 0.5 \\
learning rate & 1e-4 & 1e-4 \\
backbone lr mult. & 0.5 & 0.5 \\
lr scheduler & PolyLR & PolyLR \\
AWD scheduler & Cosine & Cosine \\
weight decay & 0.05 & 0.05 \\
optimizer & AdamW & AdamW \\
batch size & 4 & 4 \\
iterations & 40,000 & 40,000 \\
\bottomrule
\end{tabular}}
\caption{Hyperparameter settings for experiments.}
\label{tab:hyperparameters}
\end{table}
\section{Additional Experiments and Analysis}
\subsection{Hyperparameter Analysis}
\noindent \textbf{Rank Analysis.}
To determine the optimal configuration for our dual-adapter structure, we conducted an analysis to investigate the impact of the rank settings for both the Main and Sub Adapters on domain generalization performance.

First, for the synthetic-to-real scenario (Table~\ref{tab:my_model_perf}), we found that a Main Adapter rank of 32 consistently outperformed a rank of 16. With the Main Adapter's rank fixed at 32, we observed that performance peaked at an average mIoU of 68.95 when the Sub Adapter's rank was 4. However, increasing the rank further to 8 or higher led to a noticeable degradation in performance. This finding is consistent with our hypothesis that the Sub Adapter, which modifies the VFM's major directions, requires minimal and careful adjustments. Therefore, we adopted the (32, 4) rank configuration for the synthetic-to-real experiments (e.g., Table~\ref{tab: g2cbm}, ~\ref{tab: decomposition}).

We extended this analysis to the real-to-real generalization scenario (Table~\ref{tab:my_model_perf_r}) as well. In this setting, the best performance (72.10 mIoU) was achieved with a Main Adapter rank of 32 and a Sub Adapter rank of 2. This result suggests that for the real-to-real setting, which has a smaller domain gap, an even more conservative adjustment of the VFM's major directions is beneficial. Consequently, we used the (32, 2) rank configuration for all real-to-real experiments (e.g., Table~\ref{tab: c2bm}).
\begin{table}[H]
    \centering
    \centering
\colorlet{LightGray}{gray!15}
\resizebox{1\columnwidth}{!}{
{\begin{tabular}{c|c|c|c|c|c|c|c|c}
\toprule
\multicolumn{9}{c}{\textmd{\textbf{Synthetic-to-Real Generalization}}} \\ \midrule
\multicolumn{2}{c|}{rank} & \multicolumn{2}{c|}{lr} & \multirow{2}{*}{Params.} & \multicolumn{4}{c}{\textmd{\textit{Trained on GTAV}}} \\
main & sub & main & sub & & $\rightarrow$Citys. & $\rightarrow$BDD & $\rightarrow$Map. & Avg. \\
\midrule
32 & 2 & 1e-4 & 5e-5 & 13.4M & 72.03 & 60.75 & \underline{72.06} & \underline{68.28} \\
32 & 4 & 1e-4 & 5e-5 & 14.2M & \textbf{73.01} & \textbf{61.77} & \textbf{72.07} & \textbf{68.95} \\
32 & 8 & 1e-4 & 5e-5 & 15.7M & \underline{72.83} & 61.13 & 70.81 & 68.26 \\
32 & 16 & 1e-4 & 5e-5 & 18.9M & 71.20 & 61.16 & 70.87 & 67.74 \\
32 & 32 & 1e-4 & 5e-5 & 25.2M & 72.13 & 60.83 & 69.87 & 67.61 \\
16 & 2 & 1e-4 & 5e-5 & 7.1M & 71.67 & 61.27 & 70.84 & 67.93 \\
16 & 4 & 1e-4 & 5e-5 & 7.9M & 71.74 & \underline{61.54} & 71.25 & 68.18 \\
16 & 8 & 1e-4 & 5e-5 & 9.4M & 71.40 & 60.56 & 70.41 & 67.46 \\
16 & 16 & 1e-4 & 5e-5 & 12.6M & 71.51 & 60.48 & 70.37 & 67.45 \\
\bottomrule
\end{tabular}}
}
\caption{Domain generalization results (mIoU \%) for RecycleLoRA with varying rank configurations for its Main and Sub Adapters, under the synthetic-to-real setting (G$\rightarrow$\{C, B, M\}). \textbf{Bold} and \underline{underlined} indicate best and second-best results.}
\label{tab:my_model_perf}
\end{table}

\begin{table}[H]
    \centering
    \centering
\colorlet{LightGray}{gray!15}
\resizebox{0.9\columnwidth}{!}{
{\begin{tabular}{c|c|c|c|c|c|c|c}
\toprule
\multicolumn{8}{c}{\textmd{\textbf{Real-to-Real Generalization}}} \\ \midrule
\multicolumn{2}{c|}{rank} & \multicolumn{2}{c|}{lr} & \multirow{2}{*}{Params.} & \multicolumn{3}{c}{\textmd{\textit{Trained on Cityscapes}}} \\
main & sub & main & sub & & $\rightarrow$BDD & $\rightarrow$Map. & Avg. \\
\midrule
32 & 2 & 1e-4 & 5e-5 & 13.4M & \underline{66.65} & \textbf{78.14} & \textbf{72.10} \\
32 & 4 & 1e-4 & 5e-5 & 14.2M & \textbf{66.76} & \underline{76.64} & \underline{71.70} \\
\bottomrule
\end{tabular}}
}
\caption{Domain generalization results (mIoU \%) for RecycleLoRA with varying rank configurations for its Main and Sub Adapters, under the real-to-real setting (C$\rightarrow$\{B, M\}). \textbf{Bold} and \underline{underlined} indicate best and second-best results.}
\label{tab:my_model_perf_r}
\end{table}

\noindent \textbf{Learning rate Analysis.}
We conduct an analysis to determine the optimal learning rate for the Sub Adapter and validate our design choice of using a different learning rate from the Main Adapter. As shown in Table~\ref{tab:my_model_perf_lr}, we fixed the Main Adapter's learning rate to 1e-4 and varied the Sub Adapter's learning rate.
The results demonstrate that the best performance is achieved when the Sub Adapter's learning rate is set to 5e-5, half that of the Main Adapter, achieving an average mIoU of 68.95. Setting the learning rate for the Sub Adapter to be either the same as the Main Adapter (1e-4) or excessively low (1e-5) resulted in a performance drop. This empirical evidence supports our strategy of applying a carefully tuned, lower learning rate to the Sub Adapter. This differentiation is crucial for enabling the complementary learning process between the two adapters, validating our overall design.

\begin{table}[H]
    \centering
    \centering
\colorlet{LightGray}{gray!15}
\resizebox{1\columnwidth}{!}{
{\begin{tabular}{c|c|c|c|c|c|c|c|c}
\toprule
\multicolumn{9}{c}{\textmd{\textbf{Synthetic-to-Real Generalization}}} \\ \midrule
\multicolumn{2}{c|}{rank} & \multicolumn{2}{c|}{lr} & \multirow{2}{*}{Params.} & \multicolumn{4}{c}{\textmd{\textit{Trained on GTAV}}} \\
main & sub & main & sub & & $\rightarrow$Citys. & $\rightarrow$BDD & $\rightarrow$Map. & Avg. \\
\midrule
32 & 4 & 1e-4 & 1e-4 & 14.2M & 72.23 & 60.50 & \underline{70.60} & 67.78 \\
32 & 4 & 1e-4 & 5e-5 & 14.2M & \textbf{73.01} & \textbf{61.77} & \textbf{72.07} & \textbf{68.95} \\
32 & 4 & 1e-4 & 1e-5 & 14.2M & \underline{72.55} & \underline{61.10} & 70.55 & \underline{68.07} \\
\bottomrule
\end{tabular}}
}
\caption{Domain generalization results (mIoU \%) for RecycleLoRA with varying learning rate configurations for its Main and Sub Adapters, under the synthetic-to-real setting (G$\rightarrow$\{C, B, M\}). \textbf{Bold} and \underline{underlined} indicate best and second-best results.}
\label{tab:my_model_perf_lr}

\end{table}

\subsection{Ablation on Dual-Adapter Initialization}
To assess the impact of the initialization strategy on our dual-adapter framework, we conducted an ablation study comparing our RRQR-based approach with other representative initialization methods. We compare against Kaiming uniform initialization, a standard method that does not leverage the pre-trained weight structure, and SVD-based initialization, which utilizes subspace decomposition as seen in prior work such as SoMA~\cite{SoMA} or PiSSA~\cite{Pissa}. Interestingly, as presented in Table~\ref{tab: decomposition}, the other initialization methods did not synergize with the dual-adapter structure and instead exhibited performance degradation. Specifically, the dual-adapter with Kaiming initialization scored 66.31 mIoU, which is lower than the 66.89 mIoU of standard LoRA~\cite{LoRA} (single adapter) presented in Table~\ref{tab: g2cbm}. Similarly, the SVD-based initialization achieved only 67.23 mIoU, underperforming SoMA (single adapter), which scored 68.27 mIoU as shown in Table~\ref{tab: g2cbm}. In contrast, our proposed RRQR-based initialization achieves an average mIoU of 68.95, significantly outperforming both alternatives. These results underscore the importance of the initialization method and suggest that our proposed RRQR-based strategy is a more effective choice for the proposed dual-adapter structure.

\begin{table}[H]
    \vspace{-0.5mm}
    \centering
    \centering
\colorlet{LightGray}{gray!15}
\resizebox{1\columnwidth}{!}{
{\begin{tabular}{l|l|c|c|c|c}
\toprule
\multicolumn{6}{c}{\textmd{\textbf{Synthetic-to-Real Generalization}}} \\ \midrule
\multirow{2}{4em}{Initialization} & \multirow{2}{3em}{Backbone} &
\multicolumn{4}{c}{\textmd{\textit{Trained on GTAV}}} \\
 & & $\rightarrow$Citys. & $\rightarrow$BDD & $\rightarrow$Map. & Avg. \\ 
\midrule
Kaiming unif. & DINOv2-L & 68.96 & \underline{60.57} & 69.41 & 66.31 \\
SVD & DINOv2-L & \underline{70.40} & 60.35 & \underline{70.93} & \underline{67.23} \\
RRQR & DINOv2-L &\textbf{73.01} & \textbf{61.77} & \textbf{72.07} & \textbf{68.95} \\
\bottomrule
\end{tabular}}
}
\caption{Domain generalization results (mIoU \%) for the dual-adapter framework with different initialization strategies, under the synthetic-to-real setting (G\(\rightarrow\)\{C, B, M\}). \textbf{Bold} and \underline{underlined} indicate best and second-best results.}
\label{tab: decomposition}
    \vspace{-0.5mm}
\end{table}

\section{Limitations and Future Works}
While RecycleLoRA demonstrates robust state-of-the-art performance, we identify several avenues for future research that could further advance its capabilities and address its current limitations.

Further optimization of RecycleLoRA could be achieved by developing systematic methods for tuning hyperparameters, such as the ranks and learning rates of the dual adapters. The current configuration was determined empirically, and creating automated search strategies would enhance the practicality and replicability of our approach.

The scope of our work, currently focused on Domain Generalized Semantic Segmentation, could also be broadened. The core principle of recycling pre-trained knowledge through subspace analysis is likely applicable to other downstream tasks that require efficient foundation model fine-tuning, such as object detection, video analysis, and medical image segmentation. Investigating the effectiveness of our approach in these diverse contexts presents a logical direction for future work.

Another promising research direction involves revisiting our methodology's binary partitioning of the VFM's subspace into ``major" and ``minor" directions, which currently omits the intermediate directions. This simplification is based on the hypothesis that the most and least dominant directions are most critical for balancing knowledge preservation and new feature acquisition. However, the potential contribution of these intermediate directions remains unexplored. Future research could investigate a more nuanced allocation of the entire subspace spectrum, perhaps through a third adapter or a soft-weighting scheme that utilizes all ranked directions, which may unlock further performance gains.

\section{Qualitative Results}
\begin{figure*}[h]
  \centering
  \includegraphics[width=\textwidth]{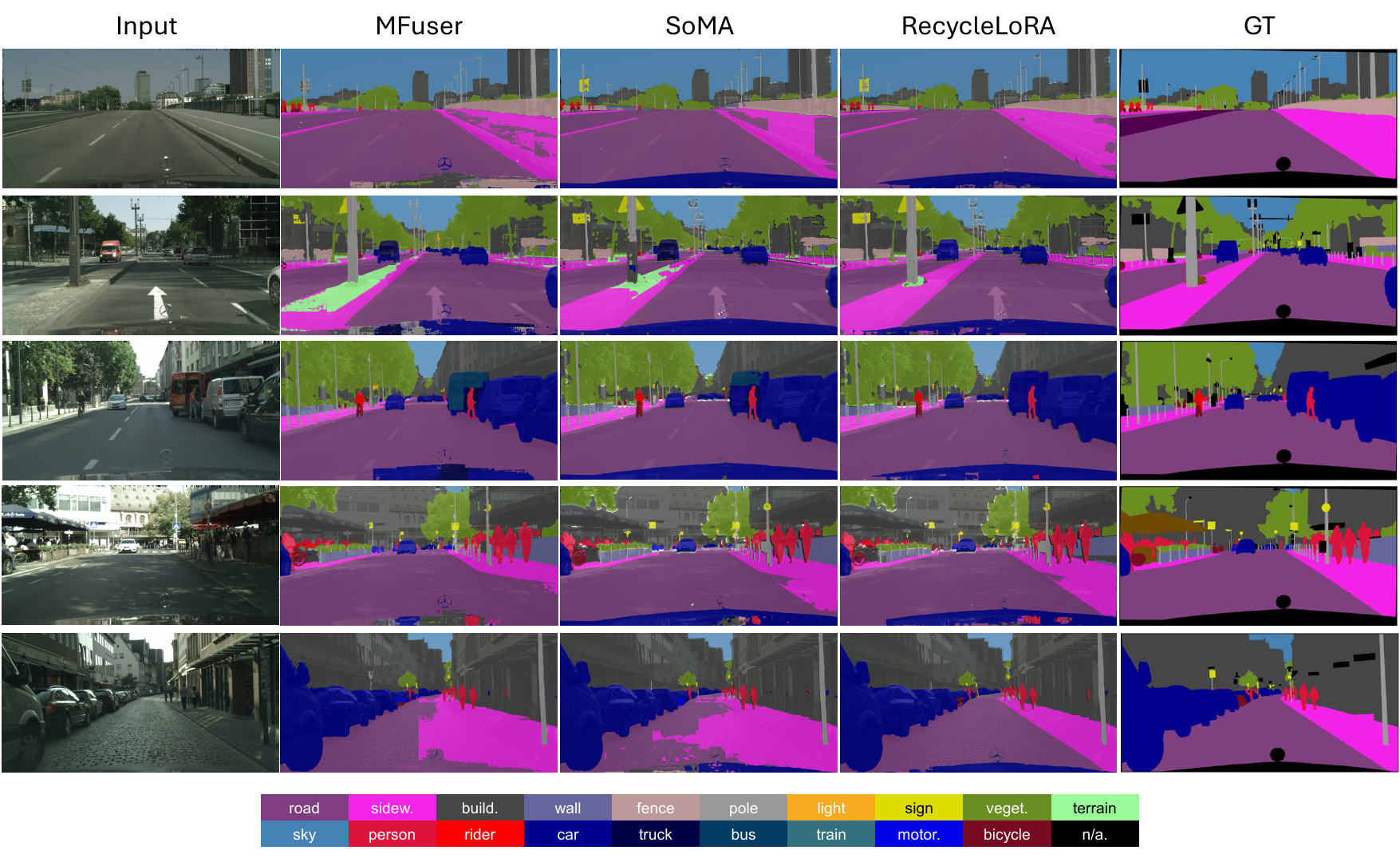}
  \caption{
    Qualitative comparison of semantic segmentation on the Cityscapes. All models were trained on the GTAV.
  }
  \label{fig:citys_viz}
\end{figure*}

\begin{figure*}[h]
  \centering
  \includegraphics[width=\textwidth]{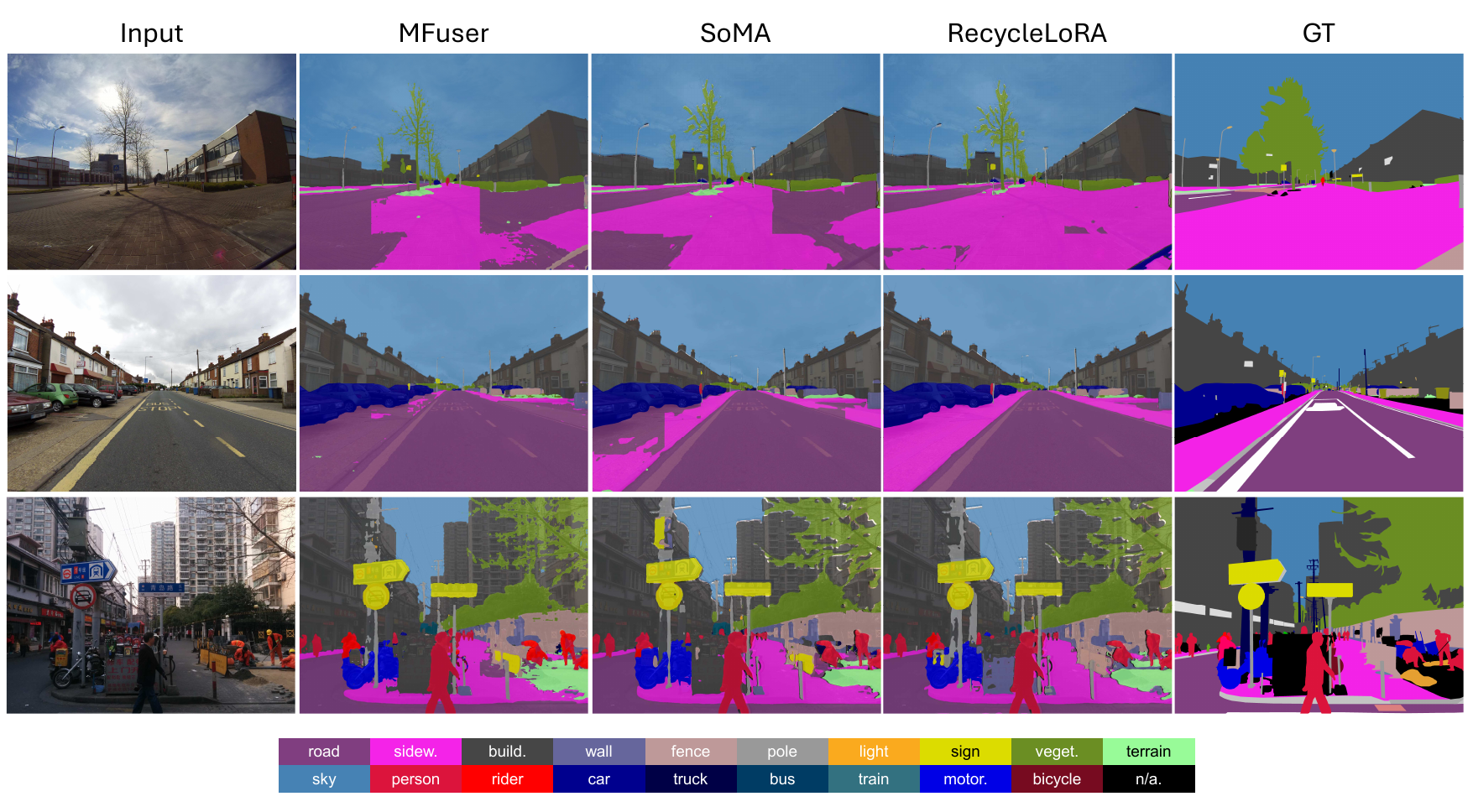}
  \caption{
    Qualitative comparison of semantic segmentation on the Mapillary. All models were trained on the GTAV.
  }
  \label{fig:map_viz}
\end{figure*}

Figures~\ref{fig:citys_viz} and~\ref{fig:map_viz} present qualitative comparisons against other state-of-the-art methods. These visualizations highlight that RecycleLoRA generates more accurate and detailed segmentation maps, which are more closely aligned with the ground truth.

\end{document}